\documentclass[journal]{IEEEtran}
\usepackage{amsmath,epsfig}
\graphicspath{{./}{KRDU_figure/}}
\usepackage{multirow}
\usepackage{graphicx,epsfig, epstopdf,subfigure}
\usepackage{algorithm, algorithmic,url}
  \usepackage{mathrsfs,amssymb,amsmath,cite, color,bm,tabu,comment}
  \usepackage[table]{xcolor}

\begin{document}
%
\title{Vision Recognition using Discriminant Sparse Optimization Learning}
%
%
%

\author{Qingxiang Feng, 
        Yicong~Zhou,~\IEEEmembership{Senior~Member,~IEEE,}
\thanks{This work was supported in part by the Macau Science and Technology Development Fund under Grant FDCT/016/2015/A1 and by the Research Committee at University of Macau under Grants MYRG2014-00003-FST and MYRG2016-00123-FST.
(Corresponding author is Yicong Zhou.)}\\
\thanks{All authors are with the Department of Computer and Information Science, University
of Macau, Macau 999078, China
(e-mail: fengqx1988@gmail.com; yicongzhou@umac.mo).}
}

\maketitle

\begin{abstract}

To better select the correct training sample and obtain the robust representation of the query sample, this paper proposes a discriminant-based sparse optimization learning model. This learning model integrates discriminant and sparsity together. Based on this model, we then propose a classifier called locality-based discriminant sparse representation (LDSR). Because discriminant can help to increase the difference of samples in different classes and to decrease the difference of samples within the same class, LDSR can obtain better sparse coefficients and constitute a better sparse representation for classification. In order to take advantages of kernel techniques, discriminant and sparsity, we further propose a nonlinear classifier called kernel locality-based discriminant sparse representation (KLDSR).  Experiments on several well-known databases prove that the performance of LDSR and KLDSR is better than that of several state-of-the-art methods including deep learning based methods.
\end{abstract}

\begin{IEEEkeywords}
Vision Recognition, Image Classification, Object Recognition, Sparse Representation based Classification, Kernel Sparse Representation Classification.
\end{IEEEkeywords}

\IEEEpeerreviewmaketitle

\section{Introduction}

With the rapidly development of imaging techniques, image classification and recognition tasks (e. g. face recognition, object recognition, action recognition etc.)  attract more and more attention. The main steps of image classification tasks include preprocessing, feature selection and classifiers. In the past few decades, a lot of researchers have worked on these three parts and have made significant improvements. Recently, many researchers pay attention to the second part (feature selection) because deep learning techniques \cite{Imagenet,NAC} can obtain the quite good feature. However, each technique may have its own bottlenecks. Although deep learning obtains quite good performance, the bottleneck of deep learning will appear sooner or later. We need the help of other techniques (e. g. classifier) to improve the final classification performance. We give an example to explain it.
\begin{quotation}
\noindent
   Example 1: Suppose that there is an examination paper that includes two parts: feature and classifier. The full mark of feature part is 80 points while the full mark of classifier part is 20 points. If an examinee only answers the first part, he/she can't get the point more than 80.
\end{quotation}
Example 1 shows that although deep learning is important, other techniques (e. g. classifier) also need to be improved.
Generally speaking, classifiers include two categories \cite{DefenseNN}\cite{MLPP}: parametric-based methods and non-parametric-based methods. The parametric-based methods (e.g., SVM \cite{SVM}) focus on learning the parameters of a hypothesis classification model from the training data \cite{ProCRC}. Then, these methods predict the class labels of unknown data using the learned parametric model. On the contrary, the non-parametric-based methods directly obtain the class labels of unknown data. The well-known non-parametric-based method is the sparse representation classification (SRC) \cite{SRC}. Compared to other classifiers, SRC and its improved versions obtain better performance for image classification. The sparsity tries to use only the correct training samples (samples have the same label of the testing sample) to constitute a robust representation for classification. However, sparsity doesn't distinguish correct training samples and incorrect training samples when they have a small difference\\
Inspired by linear discriminant analysis (LDA) \cite{pcalda,ULDA}, we know that discriminant can help to increase the difference of samples in different classes and to decrease the difference of samples within the same class. If we increase the discriminant into the optimization learning process of sparsity, we can better select the correct training samples and constitute the robust representation of the query (testing) sample. Based on that, this paper proposes a discriminant-based sparse optimization learning model that integrates the discriminant and sparsity together. Then, we further describe the detail solution procedures of the proposed discriminant-based sparse optimization learning model. Moreover, we propose a classifier called locality-based discriminant sparse representation (LDSR). In order to take advantages of the nonlinear high-dimensional feature, we propose another classifier called kernel locality-based discriminant sparse representation (KLDSR). KLDSR uses the kernel techniques \cite{KSRC}\cite{KRDU} to map the original linear feature to a high-dimensional nonlinear feature for classification. The effectiveness of the proposed classifiers are assessed on several visual classification tasks, which include face recognition on the LFW database \cite{LFW}; handwritten digit recognition on the MNIST database \cite{MNIST} and USPS dataset\cite{USPS}, flowers recognition on the Oxford 102 Flowers database \cite{Oxford102}; birds recognition on the Caltech-UCSD Birds (CUB200-2011) database \cite{CUB200}; object recognition on the Caltech 256 object databases \cite{Caltech256}; and Millions level-based image classification on the ImageNet Large Scale Visual Recognition Challenge (ILSVRC) 2012 dataset \cite{ImageNet2}.\\

\begin{figure*} [t]
\begin{center}
\subfigure{\includegraphics[width=6.2in]{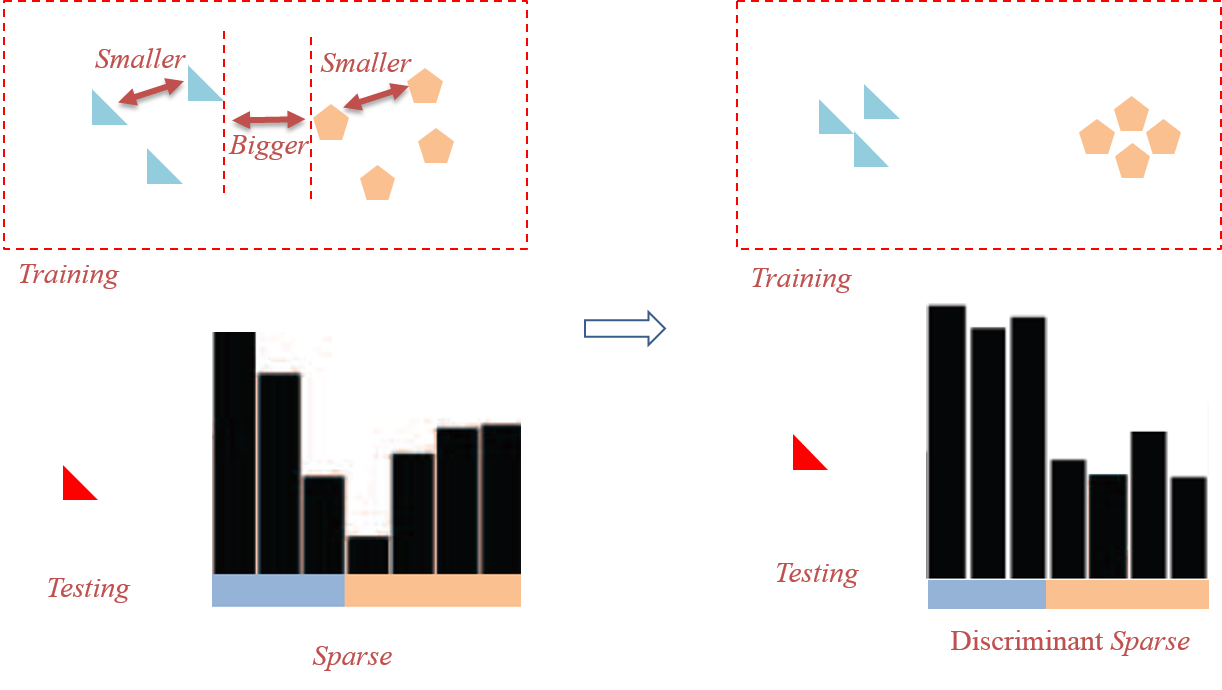}}
\end{center}
\vspace{-0.05in} \caption{ The main idea of the proposed method. Discriminant can help to increase the difference of samples in different classes and decrease the difference of samples in the same class. Given a testing sample and several training samples. If we increase the discriminant in the optimization process of sparsity, the discriminant-sparse-based coefficients in the same class would have smaller difference, and the discriminant-sparse-based coefficients in different class would have bigger difference, compared to sparse-based coefficients.} \label{fig1}
\end{figure*}
~~The main contributions of this paper are as follows:
\begin{itemize}
  \item Motivated by LDA and SRC, we propose a discriminant-based sparse optimization learning model that integrates the discriminant and sparsity together. The detail solution procedures are also given.
  \item Using the proposed discriminant-based optimization learning model, we propose a classifier called locality-based discriminant sparse representation (LDSR).
  \item Motivated by kernel techniques, we further propose kernel locality-based discriminant sparse representation (KLDSR) classifier.
  \item We evaluate the proposed LDSR and KLDSR on several well-known databases including the million-level database: ImageNet.
\end{itemize}

\section{Related Work}

This section reviews the sparse-based methods and linear discriminant analysis (LDA). Several notations are described in Table \ref{t_NotationSummary}.

\subsection{Sparse-based Methods}
\emph{Wright et al.} proposed sparse representation classification (SRC) in 2009. It represents the testing sample using the linear combination of training samples of all classes. SRC solves $L_1$-minimization optimization problem as
\begin{eqnarray}
 ||\alpha |{|_1}{\kern 1pt} {\kern 1pt} {\kern 1pt} {\kern 1pt} {\kern 1pt} {\kern 1pt} {\kern 1pt} {\kern 1pt} {\kern 1pt} {\kern 1pt} s.t.{\kern 1pt} {\kern 1pt} {\kern 1pt} X\alpha  = x
\end{eqnarray}
To reduce the computation cost and obtain the better representation, \emph{Zhang et al.} proposed the collaborative representation based classification (CRC). In CRC, the authors argued that the collaborative representation should be better than the linear combination of training samples. CRC solves the $L_2$-minimization optimization problem as
\begin{eqnarray}
||\alpha |{|_2}{\kern 1pt} {\kern 1pt} {\kern 1pt} {\kern 1pt} {\kern 1pt} {\kern 1pt} {\kern 1pt} {\kern 1pt} {\kern 1pt} {\kern 1pt} s.t.{\kern 1pt} {\kern 1pt} {\kern 1pt} X\alpha  = x
\end{eqnarray}
They are two typical sparse-based methods. A lot improved version of SRC/CRC have been proposed for various visual recognition tasks. They include the probabilistic-based sparse methods \cite{ProCRC}, discriminant-based dictionary learning \cite{FDDL,FDDL1}, kernel-based sparse methods \cite{KSRC}\cite{KCSR}, Gabor-based sparse methods \cite{GSRC} and many others \cite{SVDL, TPSR}.

\subsection{Linear discriminant analysis (LDA)}

LDA is a typical machine learning method. It aims at obtaining the largest mean differences between the desired classes. Mathematically, LDA is to maximize the Fisher-ratio criterion \cite{fisher1936use} as $\frac{{|{O_b}|}}{{|{O_w}|}}$ .  ${O_w}$ is the within-class scatter matrix, calculated by
\begin{eqnarray} \label{eq_LDAow}
O_w^{} = \sum\limits_{c = 1}^M {\sum\limits_{i = 1}^{{N_c}} {(x_i^c - {m_c}){{(x_i^c - {m_c})}^T}} }
\end{eqnarray}
where $m_c$ is the mean sample of the $c^{th}$ class.  $O_b$ is between-class scatter matrix, computed by
\begin{eqnarray} \label{eq_LDAob}
O_b^{} = \sum\limits_{c = 1}^M {({m_c} - m){{({m_c} - m)}^T}}
\end{eqnarray}
where $m$ is the mean sample of all classes.

\begin{table} [t]
\caption{Notation Summary} \label{t_NotationSummary}
\begin{center}
\begin{tabular}{|l|c|}
\hline
Notation & Explanation \\
\hline\hline
$X$  & Entire training set  \\
$X_c$  & All samples of the $c^{\mbox {th}}$ class  \\
$x_i^c$  & The $i^{\mbox{th}}$ sample of the $c^{\mbox {th}}$ class  \\
$q$  & Dimension of a sample  \\
$N_c$  & Number of samples of the $c^{\mbox{th}}$ class \\
$M$  & Number of classes  \\
$x$  & Testing sample \\
\hline
\end{tabular}
\end{center}
\end{table}

\section{Discriminant-based Sparse Optimization Learning Model}

This section proposes a discriminant-based sparse optimization learning model. To better explain this learning model, we divide this section into three subsections. Section III-A gives the motivation . Section III-B proposes a discriminant-based sparse optimization learning model. Section III-C introduces the detail solution procedures of this optimization learning model.

\subsection{Motivation}
The sparsity tries to select the correct training samples (samples have the same label of the testing sample) and to constitute a robust representation of the testing sample. However, variations (e.g. lights, views, occlusions, pose, background etc.) in the collected images make it challenging to obtain a robust representation model for image classification. Inspired by LDA, we know that discriminant can help to increase the difference of samples in different classes and to decrease the difference of samples within the same class. If we increase the discriminant into the optimization process of sparsity, the correct training samples and the incorrect training samples would have a larger difference. Then, we can better select the correct training samples to constitute the robust representation of the query sample. Fig. 1 gives an example to explain the motivation. Observing Fig. 1, we know that if we increase the discriminant in the optimization learning process of sparsity, the discriminant-sparse-based coefficients in the same class would have a smaller difference, and the discriminant-sparse-based coefficients in different classes would have a bigger difference.
The next section will describe how to increase the discriminant in the optimization learning process of sparsity.

\subsection{Discriminant-based Sparse Learning Model}

Based on the sparse representation and LDA, we propose a discriminant-based sparse learning model, its objective function is defined as
\begin{eqnarray}
\mathop {\min }\limits_\alpha  ||x - X\alpha ||_2^2 + \lambda ||\alpha ||_2^2 + \eta {S_w}+\gamma {S_b}
\end{eqnarray}
where $S_w$ means the difference of the samples within the same class. $S_b$ denotes the difference of the samples in different classes.

\subsubsection{How to obtain $S_w$}

In SRC, each class has a sparse representation. To minimize the difference of the samples within the same class, we try to minimize the distance between each samples and the corresponding class's sparse representation. Specifically, $S_w$ can be computed by
\begin{eqnarray} \label{eq_sFRCsw}
\begin{array}{c}
S_w^{} = \sum\limits_{c = 1}^M {\sum\limits_{i = 1}^{{N_c}} {{{(x_i^c\alpha _i^c - {X_c}{\alpha _c})}^T}(x_i^c\alpha _i^c - {X_c}{\alpha _c})} } \\
 = \sum\limits_{c = 1}^M {\sum\limits_{i = 1}^{{N_c}} {||x_i^c\alpha _i^c - {X_c}{\alpha _c}||_2^2} }
\end{array}
\end{eqnarray}
where $X_c$ denotes the samples of the $c^{th}$ class, ${\alpha _c} \in {R^{{N_c} \times 1}}$ is the corresponding sparse coefficient of $X_c$ , and  $\alpha _i^c \in {R^{1 \times 1}}$ is the $i^{th}$ element of $\alpha _c$.\\

\subsubsection{How to obtain $S_b$}
To increase the difference of samples in different classes, we try to minimize the correlation of different classes. That is, we want to minimize ${({X_i}{\alpha _i})^T}({X_j}{\alpha _j})$. Motivated by \cite{ASRC}, we know that minimizing $||{X_i}{\alpha _i}|{|^2}$ is also good for sparsity. Thus, $S_b$ can be obtained by
\begin{eqnarray}
\begin{array}{c}
S_b=\sum\limits_{i = 1}^M {\sum\limits_{j = 1}^M ({{||{X_i}{\alpha _i}||^2}+ 2{({X_i}{\alpha _i})^T}({X_j}{\alpha _j})+{||{X_j}{\alpha _j}||^2}) }}\\
= \sum\limits_{i = 1}^M {\sum\limits_{j = 1}^M {||{X_i}{\alpha _i} + {X_j}{\alpha _j}|{|^2}} }
\end{array}
\end{eqnarray}

Consider $S_w$ and $S_b$, the above objective can be rewritten as
\begin{eqnarray} \label{eq_of1}
\begin{array}{c}
\mathop {\min }\limits_\alpha  ||x - X\alpha ||_2^2 + \lambda |\alpha ||_2^2 + \eta \sum\limits_{c = 1}^M {\sum\limits_{i = 1}^{{N_c}} {||x_i^c\alpha _i^c - {X_c}{\alpha _c}||_2^2} } \\
 + \gamma \sum\limits_{i = 1}^M {\sum\limits_{j = 1}^M {||{X_i}{\alpha _i} + {X_j}{\alpha _j}|{|^2}} }
\end{array}
\end{eqnarray}

\subsection{Solution Procedures of Eq. (\ref{eq_of1})}
The objective function in Eq. (\ref{eq_of1}) is differentiable. Thus, the optimal solution of Eq. (\ref{eq_of1}) can be treated as the stationary point of the objective function. Let $G(\alpha ) = \mathop {\min }\limits_\alpha  ||x - X\alpha ||_2^2 + \lambda |\alpha ||_2^2 + \eta \sum\limits_{c = 1}^M {\sum\limits_{i = 1}^{{N_c}} {||x_i^c\alpha _i^c - {X_c}{\alpha _c}||_2^2} }  + \gamma \sum\limits_{i = 1}^M {\sum\limits_{j = 1}^M {||{X_i}{\alpha _i} + {X_j}{\alpha _j}|{|^2}}} $ . The derivative of the first term of  $G(\alpha )$ is
\begin{eqnarray} \label{eq_derivative1}
\frac{{\partial (||x - X\alpha |{|^2})}}{{\partial (\alpha )}} =  - 2{X^T}(x - X\alpha )
\end{eqnarray}
The derivative of the second term of  $G(\alpha )$ is
\begin{eqnarray} \label{eq_derivative2}
\frac{{\partial (\lambda ||\alpha |{|^2})}}{{\partial (\alpha )}} = 2\lambda \alpha
\end{eqnarray}
The derivative of the third term of  $G(\alpha )$ is complex. Let  $f(\alpha ) = \eta \sum\limits_{c = 1}^M {\sum\limits_{i = 1}^{{N_c}} {||x_i^c\alpha _i^c - {X_c}{\alpha _c}||_2^2} } $, we rewrite the   $f(\alpha)$ as follows
\begin{eqnarray}
\begin{array}{l}
f(\alpha ) = \eta \sum\limits_{c = 1}^M {\sum\limits_{i = 1}^{{N_c}} {||x_i^c\alpha _i^c - {X_c}{\alpha _c}||_2^2} } \\
 = \eta \sum\limits_{c = 1}^M {\sum\limits_{i = 1}^{{N_c}} {||{{\rlap{--} X}_{c,i}}{\alpha _c}||_2^2} } \\
 = \eta \sum\limits_{c = 1}^M {\sum\limits_{i = 1}^{{N_c}} {\alpha _c^T\rlap{--} X_{c,i}^T{{\rlap{--} X}_{c,i}}{\alpha _c}} } \\
 = \eta \sum\limits_{c = 1}^M {\alpha _c^T(\sum\limits_{i = 1}^{{N_c}} {\rlap{--} X_{c,i}^T{{\rlap{--} X}_{c,i}}} } ){\alpha _c}\\
 = \eta \alpha _{}^T\left( {\begin{array}{*{20}{c}}
{\sum\limits_{i = 1}^{{N_1}} {\rlap{--} X_{1,i}^T\rlap{--} X_{1,i}^{}} }&{...}&0\\
{...}&{...}&{...}\\
0&{...}&{\sum\limits_{i = 1}^{{N_M}} {\rlap{--} X_{M,i}^T\rlap{--} X_{M,i}^{}} }
\end{array}} \right)\alpha
\end{array}
\end{eqnarray}
where $\rlap{--} X_{c,i}^{} = [\begin{array}{*{20}{c}} {x_1^c}&{...}&{x_{i - 1}^c}&0&{x_{i + 1}^c}&{...}&{x_{{N_c}}^c}
\end{array}] \in {R^{q \times {N_c}}}$, $c=1,2...M$, $i=1,2...N_c$.\\
Using the rewritten $f(\alpha)$ , the derivative  $\partial f(\alpha )/\partial (\alpha )$  can be computed as
\begin{eqnarray} \label{eq_derivative3}
\begin{array}{l}
\frac{{\partial f(\alpha )}}{{\partial (\alpha )}} = \\
\frac{{\partial \left( {\eta \alpha _{}^T\left( {\begin{array}{*{20}{c}}
{\sum\limits_{i = 1}^{{N_1}} {\rlap{--} X_{1,i}^T\rlap{--} X_{1,i}^{}} }&{...}&0\\
{...}&{...}&{...}\\
0&{...}&{\sum\limits_{i = 1}^{{N_M}} {\rlap{--} X_{M,i}^T\rlap{--} X_{M,i}^{}} }
\end{array}} \right)\alpha } \right)}}{{\partial ({\alpha _c})}}\\
 = 2\eta \left( {\begin{array}{*{20}{c}}
{\sum\limits_{i = 1}^{{N_1}} {\rlap{--} X_{1,i}^T\rlap{--} X_{1,i}^{}} }&{...}&0\\
{...}&{...}&{...}\\
0&{...}&{\sum\limits_{i = 1}^{{N_M}} {\rlap{--} X_{M,i}^T\rlap{--} X_{M,i}^{}} }
\end{array}} \right)\alpha
\end{array}
\end{eqnarray}
However, the derivative of the fourth term of $g(\alpha )$  is complex. Because $f(\alpha ) = \gamma \sum\limits_{i = 1}^M {\sum\limits_{j = 1}^M {||{X_i}{\alpha _i} + {X_j}{\alpha _j}|{|^2}} } $ doesn't explicitly include $\alpha$ , we can't compute $\partial f(\alpha )/\partial (\alpha )$ directly. To solve the problem, we firstly compute the partial derivative $\partial f(\alpha )/\partial (\alpha_c )$  , and then use all $\partial f(\alpha )/\partial (\alpha_c )$ ($c=1, ..., M$) to obtain $\partial f(\alpha )/\partial (\alpha )$ . To compute the derivative $\partial f(\alpha )/\partial (\alpha_c )$  , we rewrite the  $f(\alpha ) = \gamma \sum\limits_{i = 1}^M {\sum\limits_{j = 1}^M {||{X_i}{\alpha _i} + {X_j}{\alpha _j}|{|^2}} } $  as follows
\begin{eqnarray}
\begin{array}{l}
f(\alpha ) = \gamma (\sum\limits_{\scriptstyle i = 1,...,M\hfill\atop
\scriptstyle i \ne c\hfill} {||{X_i}{\alpha _i} + {X_c}{\alpha _c}|{|^2}} \\
 \,\,\,\,\,\,\,\,\,\,\,\,\,\,\,\,\,\,\,\,\,\,\,\, + \sum\limits_{\scriptstyle j = 1,...,M\hfill\atop
\scriptstyle j \ne c\hfill} {||{X_j}{\alpha _j} + {X_c}{\alpha _c}|{|^2}} \\
 \,\,\,\,\,\,\,\,\,\,\,\,\,\,\,\,\,\,\,\,\,\,\,\, + \sum\limits_{\scriptstyle i = 1,...,M\hfill\atop
\scriptstyle i \ne c\hfill}^{} {\sum\limits_{\scriptstyle j = 1,...,M\hfill\atop
\scriptstyle j \ne c\hfill}^{} {||{X_i}{\alpha _i} + {X_j}{\alpha _j}|{|^2}} } )\\
 \,\,\,\,\,\,\,\,\,\,\,\,\,\,\,\, = \gamma (2\sum\limits_{\scriptstyle i = 1,...,M\hfill\atop
\scriptstyle i \ne c\hfill} {||{X_i}{\alpha _i} + {X_c}{\alpha _c}|{|^2}} \\
\,\,\,\,\,\,\,\,\,\,\,\,\,\,\,\,\,\,\,\,\,\,\,\, + \sum\limits_{\scriptstyle i = 1,...,M\hfill\atop
\scriptstyle i \ne c\hfill}^{} {\sum\limits_{\scriptstyle j = 1,...,M\hfill\atop
\scriptstyle j \ne c\hfill}^{} {||{X_i}{\alpha _i} + {X_j}{\alpha _j}|{|^2}} } )
\end{array}
\end{eqnarray}
Using the rewritten $f(\alpha )$ , the partial derivative  $\partial f(\alpha )/\partial (\alpha_c )$  can be computed as
\begin{eqnarray}
\begin{array}{l}
\frac{{\partial f(\alpha )}}{{\partial ({\alpha _c})}} = \frac{{\partial \left( {\gamma \sum\limits_{i = 1}^M {\sum\limits_{j = 1}^M {||{X_i}{\alpha _i} + {X_j}{\alpha _j}|{|^2}} } } \right)}}{{\partial ({\alpha _c})}}\\
 = \gamma \frac{{\partial \left( \begin{array}{c}
2\sum\limits_{\scriptstyle i = 1,...,M\hfill\atop
\scriptstyle i \ne c\hfill}^{} {||{X_c}{\alpha _c} + {X_i}{\alpha _i}|{|^2}} \;\\
\;\;\;\; + \sum\limits_{\scriptstyle i = 1,...,M\hfill\atop
\scriptstyle i \ne c\hfill}^{} {\sum\limits_{\scriptstyle j = 1,...,M\hfill\atop
\scriptstyle j \ne c\hfill}^{} {||{X_i}{\alpha _i} + {X_j}{\alpha _j}|{|^2}} }
\end{array} \right)}}{{\partial ({\alpha _c})}}\\
 = \gamma \frac{{\partial \left( {2\sum\limits_{\scriptstyle i = 1,...,M\hfill\atop
\scriptstyle i \ne c\hfill}^{} {||{X_c}{\alpha _c} + {X_i}{\alpha _i}|{|^2}} } \right)}}{{\partial ({\alpha _c})}}\\
 = 2\gamma \frac{{\partial \left( {\sum\limits_{\scriptstyle i = 1,...,M\hfill\atop
\scriptstyle i \ne c\hfill}^{} {(2X_c^T({X_c}{\alpha _c} + {X_i}{\alpha _i}))} } \right)}}{{\partial ({\alpha _c})}}\\
 = 4\gamma X_c^T\left( {(M - 1){X_c}{\alpha _c} + \sum\limits_{\scriptstyle i = 1,...,M\hfill\atop
\scriptstyle i \ne c\hfill}^{} {{X_i}{\alpha _i}} } \right)\\
 = 4\gamma X_c^T\left( {(M - 2){X_c}{\alpha _c} + \sum\limits_{i = 1,...,M}^{} {{X_i}{\alpha _i}} } \right)\\
 = 4\gamma X_c^T\left( {(M - 2){X_c}{\alpha _c} + X\alpha } \right)
\end{array}
\end{eqnarray}
Next, we use all $\partial f(\alpha )/\partial (\alpha_c )$ ($c=1, …, M$) to compute the $\partial f(\alpha )/\partial (\alpha )$  as
\begin{eqnarray} \label{eq_derivative4}
\begin{array}{l}
\frac{{\partial f(\alpha )}}{{\partial (\alpha )}} = \left( {\begin{array}{*{20}{c}}
{\frac{{\partial f(\alpha )}}{{\partial ({\alpha _1})}}}\\
{...}\\
{\frac{{\partial f(\alpha )}}{{\partial ({\alpha _M})}}}
\end{array}} \right)\\
 = \left( {\begin{array}{*{20}{c}}
{4\gamma X_1^T\left( {(M - 2){X_1}{\alpha _1} + X\alpha } \right)}\\
{...}\\
{4\gamma X_M^T\left( {(M - 2){X_M}{\alpha _M} + X\alpha } \right)}
\end{array}} \right)\\
 = 4\gamma (M - 2)\left( {\begin{array}{*{20}{c}}
{X_1^T{X_1}}&{...}&0\\
{...}&{...}&{...}\\
0&{...}&{X_M^T{X_M}}
\end{array}} \right)\alpha + 4\gamma X_{}^T{X_{}}\alpha
\end{array}
\end{eqnarray}

Using Eqs. (\ref{eq_derivative1}), (\ref{eq_derivative2}), (\ref{eq_derivative3}) and (\ref{eq_derivative4}), we can obtain the derivative $\partial G(\alpha )/\partial (\alpha )$ as
\begin{eqnarray} \label{eq_derivativeall}
\begin{array}{l}
\frac{{\partial G(\alpha )}}{{\partial (\alpha )}} =  - 2{X^T}(x - X\alpha ) + 2\lambda \alpha  + 4\gamma X_{}^T{X_{}}\alpha \\
\;\;\;\; + 2\eta \left( {\begin{array}{*{20}{c}}
{\sum\limits_{i = 1}^{{N_1}} {\rlap{--} X_{1,i}^T\rlap{--} X_{1,i}^{}} }&{...}&0\\
{...}&{...}&{...}\\
0&{...}&{\sum\limits_{i = 1}^{{N_M}} {\rlap{--} X_{M,i}^T\rlap{--} X_{M,i}^{}} }\\
\end{array}} \right)\alpha \\
\;\;\;\; +4\gamma (M - 2)\left( {\begin{array}{*{20}{c}}
{X_1^T{X_1}}&{...}&0\\
{...}&{...}&{...}\\
0&{...}&{X_M^T{X_M}}
\end{array}} \right)\alpha
\end{array}
\end{eqnarray}

\vspace{0.0in}
\begin{algorithm*} [t]
\small
\vspace{0.0in} \caption{Locality-based Discriminant Sparse Representation} \label{al_LDSR}
\begin{description}
  \item[Inputs] The entire training set $X$ with $M$ classes ${X_c} \in {R^{q \times {L_c}}}$  for $c=1, 2, ..., M$ and a testing sample ${x} \in {R^{q \times {1}}}$.
  \item[Output] class label of $x$.
\end{description}
\begin{algorithmic}[1]
  \STATE Use $X$ and $x$ to solve discriminant-based sparse coefficient $\alpha$ by
  \[\alpha  = \left( {X_{}^T{X_{}} + \lambda I + \gamma \left( {\begin{array}{*{20}{c}}
{X_1^T{X_1}}&{...}&0\\
{...}&{...}&{...}\\
0&{...}&{X_M^T{X_M}}
\end{array}} \right)  + \eta \left( {\begin{array}{*{20}{c}}
{\sum\limits_{i = 1}^{{N_1}} {\rlap{--} X_{1,i}^T\rlap{--} X_{1,i}^{}} }&{...}&0\\
{...}&{...}&{...}\\
0&{...}&{\sum\limits_{i = 1}^{{N_M}} {\rlap{--} X_{M,i}^T\rlap{--} X_{M,i}^{}} }
\end{array}} \right)} \right)X_{}^Tx\]
  where $\rlap{--} X_{c,i}^{} = [\begin{array}{*{20}{c}} {x_1^c}&{...}&{x_{i - 1}^c}&0&{x_{i + 1}^c}&{...}&{x_{{N_c}}^c}
\end{array}] \in {R^{q \times {N_c}}}$, $c=1,2...M$, $i=1,2...N_c$.
  \STATE  Compute the sparse-based distance between the testing sample and training samples by
  $d_i^c = ||x - x_i^c\alpha _i^c||$.
Constitute the locality-based training set $Y$ by selecting $s$ samples from $X$  .
  \STATE  Use $Y$ and $x$ to solve discriminant-based sparse coefficient $\beta$ by
  \[\beta  = \left( {Y_{}^T{Y_{}} + \lambda I + \gamma \left( {\begin{array}{*{20}{c}}
{Y_1^T{Y_1}}&{...}&0\\
{...}&{...}&{...}\\
0&{...}&{Y_M^T{Y_M}}
\end{array}} \right)  + \eta \left( {\begin{array}{*{20}{c}}
{\sum\limits_{i = 1}^{{N_1}} {\rlap{--} Y_{1,i}^T\rlap{--} Y_{1,i}^{}} }&{...}&0\\
{...}&{...}&{...}\\
0&{...}&{\sum\limits_{i = 1}^{{N_M}} {\rlap{--} Y_{M,i}^T\rlap{--} Y_{M,i}^{}} }
\end{array}} \right)} \right)Y_{}^Tx\]
  where $\rlap{--} Y_{c,i}^{} = [\begin{array}{*{20}{c}} {y_1^c}&{...}&{y_{i - 1}^c}&0&{y_{i + 1}^c}&{...}&{y_{{L_c}}^c}
\end{array}] \in {R^{q \times {L_c}}}$, $c=1,2...M$, $i=1,2...L_c$, $L_c$ is number of samples of $Y_c$.
  \STATE The distance between $x$ and the $c^{th}$ class is computed by
  ${s_c} = \frac{{||x - {Y_c}{\beta _c}||}}{{||{\beta _c}||}}$. Classify the testing sample $x$ into the class with the minimization distance by \[c* = \arg \min ({s_c})\]
\end{algorithmic}

\end{algorithm*}

The solution can be obtained when the condition $\partial G(\alpha )/\partial (\alpha ) = 0$ is satisfied. That is
\begin{eqnarray} \label{eq_derivativeall0}
\begin{array}{l}
0 =  - 2{X^T}(x - X\alpha ) + 2\lambda \alpha  + 4\gamma X_{}^T{X_{}}\alpha \\
\;\;\;\; + 2\eta \left( {\begin{array}{*{20}{c}}
{\sum\limits_{i = 1}^{{N_1}} {\rlap{--} X_{1,i}^T\rlap{--} X_{1,i}^{}} }&{...}&0\\
{...}&{...}&{...}\\
0&{...}&{\sum\limits_{i = 1}^{{N_M}} {\rlap{--} X_{M,i}^T\rlap{--} X_{M,i}^{}} }\\
\end{array}} \right)\alpha \\
\;\;\;\; +4\gamma (M - 2)\left( {\begin{array}{*{20}{c}}
{X_1^T{X_1}}&{...}&0\\
{...}&{...}&{...}\\
0&{...}&{X_M^T{X_M}}
\end{array}} \right)\alpha
\end{array}
\end{eqnarray}
Solving the Eq. (\ref{eq_derivativeall0}), we can obtain the optimal solution as
\begin{eqnarray} \label{eq_solution1}
\alpha  = \left( {X_{}^T{X_{}} + \lambda I + \eta H^I + 2 \gamma (M-2) H^{II}  } \right)X_{}^Tx
\end{eqnarray}
where $H^I$ is
\begin{eqnarray}
H^I = \left( {\begin{array}{*{20}{c}}
{\sum\limits_{i = 1}^{{N_1}} {\rlap{--} X_{1,i}^T\rlap{--} X_{1,i}^{}} }&{...}&0\\
{...}&{...}&{...}\\
0&{...}&{\sum\limits_{i = 1}^{{N_M}} {\rlap{--} X_{M,i}^T\rlap{--} X_{M,i}^{}} }
\end{array}} \right)
\end{eqnarray}
and $\rlap{--} X_{c,i}^{} = [\begin{array}{*{20}{c}} {x_1^c}&{...}&{x_{i - 1}^c}&0&{x_{i + 1}^c}&{...}&{x_{{N_c}}^c}
\end{array}] \in {R^{q \times {N_c}}}$, $c=1,2...M$, $i=1,2...N_c$. $H^{II}$ is
\begin{eqnarray}
H^{II} = \left( {\begin{array}{*{20}{c}}
{X_1^T{X_1}}&{...}&0\\
{...}&{...}&{...}\\
0&{...}&{X_M^T{X_M}}
\end{array}} \right)
\end{eqnarray}

\section{Locality-based Discriminant Sparse Representation}

Using the discriminant-based sparse optimization learning model in Section III, this section proposes a classifier, called locality-based discriminant sparse representation (LDSR).
Sparse representation tries to obtain a representation constituted only by the training samples with the same label of the testing sample. However, the sparsity constraint cannot guarantee it because the samples of other classes often have the effect of constituting the representation. To solve this, we try to reserve the training samples with the same label of the testing sample, and to decrease the training samples with the different label of the testing sample. Motivated by this, we try to obtain the locality-based training set as follows.
Constitute the discriminant-based optimization learning model by Eq. (\ref{eq_of1}). Compute the sparse-based distance between the testing sample and training samples by
\begin{eqnarray}
d_i^c = ||x - x_i^c\alpha _i^c||
\end{eqnarray}
Select $s$ samples with the smallest distances to constitute the locality-based training set $Y$.
Using $Y$, we constitute the discriminant-based optimization function as
\begin{eqnarray} \label{eq_of2}
\begin{array}{c}
\mathop {\min }\limits_\beta  ||x - Y\beta ||_2^2 + \lambda ||\beta ||_2^2 + \eta \sum\limits_{c = 1}^M {\sum\limits_{i = 1}^{{L_c}} {||y_i^c\beta _i^c - {Y_c}{\beta _c}||_2^2} } \\
 + \gamma \sum\limits_{i = 1}^M {\sum\limits_{j = 1}^M {||{Y_i}{\alpha _i} + {Y_j}{\alpha _j}|{|^2}}}
\end{array}
\end{eqnarray}
where $Y_c$ means the samples in the $c^{th}$ class, $L_c$ is the number of samples of $Y_c$, $y^c_i$ is the $i^{th}$ sample of the $c^{th}$ class, $i=1,2...L_c$.\\
Solve Eq. (\ref{eq_of2}) and obtain the sparse coefficient as
\begin{eqnarray} \label{eq_solution1}
\beta  = \left( {Y_{}^T{Y_{}} + \lambda I + \eta P^I + 2 \gamma (M-2) P^{II}  } \right)Y_{}^Tx
\end{eqnarray}
where $P^I$ is
\begin{eqnarray}
P^I = \left( {\begin{array}{*{20}{c}}
{\sum\limits_{i = 1}^{{N_1}} {\rlap{--} Y_{1,i}^T\rlap{--} Y_{1,i}^{}} }&{...}&0\\
{...}&{...}&{...}\\
0&{...}&{\sum\limits_{i = 1}^{{N_M}} {\rlap{--} Y_{M,i}^T\rlap{--} Y_{M,i}^{}} }
\end{array}} \right)
\end{eqnarray}
and $\rlap{--} Y_{c,i}^{} = [\begin{array}{*{20}{c}} {x_1^c}&{...}&{x_{i - 1}^c}&0&{x_{i + 1}^c}&{...}&{x_{{L_c}}^c}
\end{array}] \in {R^{q \times {N_c}}}$, $c=1,2...M$, $i=1,2...L_c$. $P^{II}$ is
\begin{eqnarray}
P^{II} = \left( {\begin{array}{*{20}{c}}
{Y_1^T{Y_1}}&{...}&0\\
{...}&{...}&{...}\\
0&{...}&{Y_M^T{Y_M}}
\end{array}} \right)
\end{eqnarray}
Compute the distance between $x$ and the $c^{th}$ class as
\begin{eqnarray}
{s_c} = \frac{{||x - {Y_c}{\beta _c}||}}{{||{\beta _c}||}}
\end{eqnarray}
Classify the testing sample $x$ with the minimization distance by
\begin{eqnarray}
c* = \arg \min ({s_c})
\end{eqnarray}
The detail procedures of LDSR are summarized in Algorithm 1.

\vspace{0.0in}
\begin{algorithm*} 
\small
\vspace{0.0in} \caption{Kernel Locality-based Discriminant Sparse Representation} \label{al_KLDSR}
\begin{description}
  \item[Inputs] The entire training set $X$ with $M$ classes ${X_c} \in {R^{q \times {L_c}}}$  for $c=1, 2, ..., M$ and a testing sample ${x} \in {R^{q \times {1}}}$.
  \item[Output] class label of $x$.
\end{description}
\begin{algorithmic}[1]
  \STATE Constitute the kernel matrix $K$ and testing kernel vector $k(.,x)$ by
  \[K = \left[ {\begin{array}{*{20}{c}}
{k(x_1^{},x_1^{})}&{k(x_1^{},x_2^{})}&{...}&{k(x_1^{},x_L^{})}\\
{k(x_2^{},x_1^1)}&{k(x_2^{},x_2^{})}&{...}&{k(x_2^{},x_L^{})}\\
{...}&{...}&{...}&{...}\\
{k(x_L^{},x_1^{})}&{k(x_L^{},x_1^{})}&{...}&{k(x_L^{},x_L^{})}
\end{array}} \right] ~~~~~~~~~~~~~~~~~~~~~~~~
k(.,x)  = {[\begin{array}{*{20}{c}}
{k({x_1},x)}&{k({x_2},x)}&{...}&{k({x_L},x)}
\end{array}]^T} \]
  \STATE Utilize $K$ and $k(.,x)$ to solve discriminant-based sparse coefficient $\alpha$ by
  \[\alpha  = \left( {K_{}^T{K_{}} + \lambda I + \gamma \left( {\begin{array}{*{20}{c}}
{K_1^T{K_1}}&{...}&0\\
{...}&{...}&{...}\\
0&{...}&{K_M^T{K_M}}
\end{array}} \right)  + \eta \left( {\begin{array}{*{20}{c}}
{\sum\limits_{i = 1}^{{N_1}} {\rlap{--} K_{1,i}^T\rlap{--} K_{1,i}^{}} }&{...}&0\\
{...}&{...}&{...}\\
0&{...}&{\sum\limits_{i = 1}^{{N_M}} {\rlap{--} K_{M,i}^T\rlap{--} K_{M,i}^{}} }
\end{array}} \right)} \right)K_{}^Tk(.,x)\]
  where $\rlap{--} K_{c,i}^{} = {\phi{(X)}}^T[\begin{array}{*{20}{c}} {x_1^c}&{...}&{x_{i - 1}^c}&0&{x_{i + 1}^c}&{...}&{x_{{N_c}}^c}
\end{array}] $, $c=1,2...M$, $i=1,2...N_c$
  \STATE  Compute the nonlinear sparse-based distance between the testing sample and training samples by
  \[\begin{split}
d_i^c = ||\phi (x) - \phi (x_i^c)\alpha||
 = \sqrt {{{(\phi (x) - \phi (x_i^c)\alpha)}^T}(\phi (x) - \phi (x_i^c)\alpha)}
 = \sqrt {k(x,x) - 2k(x,x_i^c)\alpha + \alpha^Tk(x_i^c,x_i^c)\alpha}
\end{split}\]
Select $s$ samples as $Y$ and constitute a locality-based kernel matrix $U$ and a testing kernel vector $u(.,x)$. The calculation methods of $U$ and $u(.,x)$ are similar to those of $K$ and $k(.,x)$.
  \STATE  Use $U$ and $u(.,x)$ to solve discriminant-based sparse coefficient $\beta$ by
  \[\beta  = \left( {U_{}^T{U_{}} + \lambda I + \gamma \left( {\begin{array}{*{20}{c}}
{U_1^T{U_1}}&{...}&0\\
{...}&{...}&{...}\\
0&{...}&{U_M^T{U_M}}
\end{array}} \right)  + \eta \left( {\begin{array}{*{20}{c}}
{\sum\limits_{i = 1}^{{N_1}} {\rlap{--} U_{1,i}^T\rlap{--} U_{1,i}^{}} }&{...}&0\\
{...}&{...}&{...}\\
0&{...}&{\sum\limits_{i = 1}^{{N_M}} {\rlap{--} U_{M,i}^T\rlap{--} U_{M,i}^{}} }
\end{array}} \right)} \right)U_{}^Tu(.,x)\]
  where $\rlap{--} U_{c,i}^{} = {\phi{(Y)}}^T [\begin{array}{*{20}{c}} {y_1^c}&{...}&{y_{i - 1}^c}&0&{y_{i + 1}^c}&{...}&{y_{{L_c}}^c}
\end{array}]$, $c=1,2...M$, $i=1,2...L_c$, $L_c$ is number of samples of $Y_c$.
  \STATE The distance between $x$ and the $c^{th}$ class is computed by
  ${s_c} = \frac{{||u(.,x) - {U_c}{\beta _c}||}}{{||{\beta _c}||}}$. Classify the testing sample $x$ into the class with the minimization distance by \[c* = \arg \min ({s_c})\]
\end{algorithmic}

\end{algorithm*}

\section{Proposed KLDSR} \label{sect_III}
In this Section, we propose the kernel locality-based Discriminant Sparse Representation (KLDSR) classifier. The detail processes of KLDSR are summarized in Algorithm 2.

\subsection{Kernel trick} \label{sec_IIIA}
In order to map a linear feature to a high-dimensional nonlinear feature, the kernel techniques \cite{KSRC} have been proposed.  In this section, we utilizes the most popular Gaussian radial basis function (RBF) kernel for classification. The RBF kernel can be described as
\begin{equation}\label{eq16}
k(x,y) = \phi {(x)^T}\phi (y) = \exp ( - \frac{{||x - y||_{}^2}}{\sigma })
\end{equation}
where $x$ and $y$ denote any two original samples, $\sigma$ denotes a parameter. In kernel methods, $\phi (*)$ is unknown. We can use only $k(*,*)$ to access the feature space.

\subsection{Nonlinear objective function} \label{sec_IIIB}
We suppose that there is a nonlinear feature mapping function  $\Phi (.):{{\rm{R}}^q} \to {{\rm{R}}^Q}(q <  < Q)$. This function maps the testing sample $x$ and training set $X$  into a high-dimensional feature space as
\begin{equation}\label{eq17}
\begin{array}{l}
x \to \Phi (x)\\
X \to \Phi (X) = [\begin{array}{*{20}{c}}
{\Phi ({x_1})}&{...}&{\Phi ({x_i})}&{...}&{\Phi ({x_L})}
\end{array}]
\end{array}
\end{equation}
The nonlinear objective function can be described as
\begin{eqnarray} \label{eq_nof1}
\begin{array}{c}
\mathop {\min }\limits_\alpha  ||\phi {(x)} - \phi {(X)}\alpha ||_2^2 + \lambda ||\alpha ||_2^2 \\
+ \eta \sum\limits_{c = 1}^M {\sum\limits_{i = 1}^{{N_c}} {||\phi {(x_i^c)}\alpha _i^c - \phi {{(X_c)}}{\alpha _c}||_2^2} } \\
 + \gamma \sum\limits_{i = 1}^M {\sum\limits_{j = 1}^M {||\phi {{(X_i)}}{\alpha _i} + \phi {{(X_j)}}{\alpha _j}|{|^2}} }
\end{array}
\end{eqnarray}
Because the dimension of nonlinear information is quite high, it is difficult to solve the above formula. We use the kernel trick and obtain the following objective function.
\begin{eqnarray} \label{eq_nof2}
\begin{array}{c}
\mathop {\min }\limits_\alpha  ||{\phi {(X)}}^T \phi {(x)} - {\phi {(X)}}^T \phi {(X)}\alpha ||_2^2 + \lambda ||\alpha ||_2^2 \\
+ \eta \sum\limits_{c = 1}^M {\sum\limits_{i = 1}^{{N_c}} {||{\phi {(X)}}^T \phi {(x_i^c)}\alpha _i^c - {\phi {(X)}}^T \phi {{(X_c)}}{\alpha _c}||_2^2} } \\
 + \gamma \sum\limits_{i = 1}^M {\sum\limits_{j = 1}^M {||{\phi {(X)}}^T \phi {{(X_i)}}{\alpha _i} + {\phi {(X)}}^T \phi {{(X_j)}}{\alpha _j}|{|^2}} }
\end{array}
\end{eqnarray}
Using the kernel technique, we get a kernel matrix $K = \Phi {(X)^T}\Phi (X)$ as
\begin{equation}\label{eq_kr1}
K = \left[ {\begin{array}{*{20}{c}}
{k(x_1^{},x_1^{})}&{k(x_1^{},x_2^{})}&{...}&{k(x_1^{},x_L^{})}\\
{k(x_2^{},x_1^1)}&{k(x_2^{},x_2^{})}&{...}&{k(x_2^{},x_L^{})}\\
{...}&{...}&{...}&{...}\\
{k(x_L^{},x_1^{})}&{k(x_L^{},x_1^{})}&{...}&{k(x_L^{},x_L^{})}
\end{array}} \right]
\end{equation}
and $K_c = \Phi {(X)^T}\Phi (X_c)$ is the corresponding columns ($X_c$) of K,
and $K^i_c = \Phi {(X)^T}\Phi (x^i_c)$
\begin{equation}\label{eq_kr2}
\begin{split}
K^i_c  = {[\begin{array}{*{20}{c}}
{k({x_1},x^i_c)}&{k({x_2},x^i_c)}&{...}&{k({x_L},x^i_c)}
\end{array}]^T}
\end{split}
\end{equation}
and a testing vector $k(.,x) = \Phi {(Y)^T}\Phi (x)$ as
\begin{equation}\label{eq_kr3}
\begin{split}
k(.,x) =& \phi {(X)^T}\phi (x)\\
 =& {[\begin{array}{*{20}{c}}
{k({x_1},x)}&{k({x_2},x)}&{...}&{k({x_L},x)}
\end{array}]^T}
\end{split}
\end{equation}
Consider Eqs. (\ref{eq_kr1})-(\ref{eq_kr3}), the nonlinear objective function can be rewritten as
\begin{eqnarray} \label{eq_nof3}
\begin{array}{c}
\mathop {\min }\limits_\alpha  ||k(.,x) - K\alpha ||_2^2 + \lambda ||\alpha ||_2^2 \\
+ \eta \sum\limits_{c = 1}^M {\sum\limits_{i = 1}^{{N_c}} {||K_c^i\alpha _i^c - K_c{\alpha _c}||_2^2} } \\
 + \gamma \sum\limits_{i = 1}^M {\sum\limits_{j = 1}^M {||K_i{\alpha _i} + K_j{\alpha _j}|{|^2}} }
\end{array}
\end{eqnarray}
Solving Eq. (\ref{eq_nof3}), we can obtain the optimal solution as
\begin{eqnarray} \label{eq_solution1}
\alpha  = \left( {K_{}^T{K_{}} + \lambda I + \eta A^I + 2 \gamma (M-2) A^{II}  } \right)K_{}^Tx
\end{eqnarray}
where $A^I$ is
\begin{eqnarray}
A^I = \left( {\begin{array}{*{20}{c}}
{\sum\limits_{i = 1}^{{N_1}} {\rlap{--} K_{1,i}^T\rlap{--} K_{1,i}^{}} }&{...}&0\\
{...}&{...}&{...}\\
0&{...}&{\sum\limits_{i = 1}^{{N_M}} {\rlap{--} K_{M,i}^T\rlap{--} K_{M,i}^{}} }
\end{array}} \right)
\end{eqnarray}
and $\rlap{--} K_{c,i}^{} = {\phi{(X)}}^T[\begin{array}{*{20}{c}} {x_1^c}&{...}&{x_{i - 1}^c}&0&{x_{i + 1}^c}&{...}&{x_{{N_c}}^c}
\end{array}] $, $c=1,2...M$, $i=1,2...N_c$, and $A^{II}$ is
\begin{eqnarray}
A^{II} = \left( {\begin{array}{*{20}{c}}
{K_1^T{K_1}}&{...}&0\\
{...}&{...}&{...}\\
0&{...}&{K_M^T{K_M}}
\end{array}} \right)
\end{eqnarray}

Compute the sparse-based distance between the testing sample and training samples by
\begin{equation}
\begin{split}
d_i^c =& ||\phi (x) - \phi (x_i^c)\alpha||\\
 =& \sqrt {{{(\phi (x) - \phi (x_i^c)\alpha)}^T}(\phi (x) - \phi (x_i^c)\alpha)} \\
 =& \sqrt {k(x,x) - 2k(x,x_i^c)\alpha + \alpha^Tk(x_i^c,x_i^c)\alpha}
\end{split}
\end{equation}
Select $s$ samples as $Y$ and constitute a locality-based kernel matrix $U$ and a testing kernel vector $u(.,x)$. The calculation methods of $U$ and $u(.,x)$ are similar to those of $K$ and $k(.,x)$.
Using $U$ and $u(.,x)$, the nonlinear-discriminant-based optimization function is described as
\begin{eqnarray} \label{eq_nof4}
\begin{array}{c}
\mathop {\min }\limits_\beta  ||u(.,x) - U\beta ||_2^2 + \lambda ||\beta ||_2^2 \\
+ \eta \sum\limits_{c = 1}^M {\sum\limits_{i = 1}^{{L_c}} {||U_i^c\beta _i^c - {U_c}{\beta _c}||_2^2} } \\
 + \gamma \sum\limits_{i = 1}^M {\sum\limits_{j = 1}^M {||{U_i}{\alpha _i} + {U_j}{\alpha _j}|{|^2}}}
\end{array}
\end{eqnarray}
where $U_c$ indicates a sub-kernel-matrix constituted by samples in the $c^{th}$ class, $L_c$ is the number of samples of $U_c$.\\
Solving Eq. (\ref{eq_nof4}), we can obtain the optimal solution as
\begin{eqnarray} \label{eq_solution1}
\alpha  = \left( {U_{}^T{U_{}} + \lambda I + \eta B^I + 2 \gamma (M-2) B^{II}  } \right)U_{}^Tx
\end{eqnarray}
where $B^I$ is
\begin{eqnarray}
B^I = \left( {\begin{array}{*{20}{c}}
{\sum\limits_{i = 1}^{{N_1}} {\rlap{--} U_{1,i}^T\rlap{--} U_{1,i}^{}} }&{...}&0\\
{...}&{...}&{...}\\
0&{...}&{\sum\limits_{i = 1}^{{N_M}} {\rlap{--} U_{M,i}^T\rlap{--} U_{M,i}^{}} }
\end{array}} \right)
\end{eqnarray}
and $\rlap{--} U_{c,i}^{} = {\phi{(Y)}}^T[\begin{array}{*{20}{c}} {y_1^c}&{...}&{y_{i - 1}^c}&0&{y_{i + 1}^c}&{...}&{y_{{L_c}}^c}
\end{array}] $, $c=1,2...M$, $i=1,2...L_c$, and $B^{II}$ is
\begin{eqnarray}
B^{II} = \left( {\begin{array}{*{20}{c}}
{U_1^T{U_1}}&{...}&0\\
{...}&{...}&{...}\\
0&{...}&{U_M^T{U_M}}
\end{array}} \right)
\end{eqnarray}
Compute the distance between $x$ and the $c^{th}$ class as
\begin{eqnarray}
{s_c} = \frac{{||u(.,x) - {U_c}{\beta _c}||}}{{||{\beta _c}||}}
\end{eqnarray}
Classify the testing sample $x$ with the minimization distance by
\begin{eqnarray}
c* = \arg \min ({s_c})
\end{eqnarray}
The detail procedures of KLDSR are summarized in Algorithm 2.

\section{Experimental Results}

The performance of the proposed LDSR and KLDSR classifiers is evaluated on several vision recognition databases: face recognition on the LFW database \cite{LFW}; handwritten digit recognition on the MNIST database \cite{MNIST} and USPS dataset\cite{USPS}, flowers recognition on the Oxford 102 Flowers database \cite{Oxford102}; birds recognition on the Caltech-UCSD Birds (CUB200-2011) database \cite{CUB200}; object recognition on the Caltech 256 object databases \cite{Caltech256}; and Millions level-based image classification on the ImageNet Large Scale Visual Recognition Challenge (ILSVRC) 2012 dataset \cite{ImageNet2}.
\subsection{Face recognition}

LFW face database is used in this experiment.
Following the protocol in \cite{MCT}, we apply 158 subjects that have no less than ten samples for evaluation. The experiment here is set as follows :
For each person, 5 samples are randomly selected to form the training set, while other 2 samples are exploited for testing.
The SRC \cite{SRC}, SVM \cite{SVM}, FDDL \cite{FDDL}, MCT \cite {MCT}, NSC \cite{LRC}, ProCRC \cite{ProCRC} and CRC \cite{CRC} algorithms are chosen for comparison.
Table \ref{t_lfw} illustrates the comparison results of all methods. Compared to these exsiting methods, KLDSR has more than 1\% improvement.

\begin{table} [t]
\caption{Recognition rates (RR) of several classifiers on LFW face database} \label{t_lfw}
\setlength{\tabcolsep}{10.75pt}
\setlength{\extrarowheight}{3.75pt}
\begin{center}
\begin{tabular}{|l|c|}
\hline
Classifier & Accuracy (\%)\\
\hline\hline
SVM  & 43.30  \\
NSC  & 43.80  \\
SRC  & 44.10  \\
KSRC & 44.80 \\
CRC  & 44.30 \\
FDDL  & 42.00  \\
MCT  & 44.90  \\
NDSR & 45.70 \\
ProCRC  & 45.80 \\

\hline
\bf LDSR  & \bf 46.40 \\
\bf KLDSR  & \bf 46.80 \\
\hline
\end{tabular}
\end{center}
\end{table}

\subsection{Handwritten Digit Recognition}

This section uses two databases (MNIST dataset and USPS dataset) to evaluate the performance of proposed DSR for handwritten digit recognition. These two databases are described as follows.

MNIST dataset: This handwritten digit dataset has 70000 images: 60,000 handwritten digit images are the training set and the rest 10,000 handwritten digit images are testing set. There are 10 digits (0-9) in total. Each handwritten digit image is with the size of $28\times 28$.

USPS dataset: This handwritten digit dataset has 9306 images, the training set includes 7291 images and the rest 2007 handwritten digit images are testing set. It also has ten digits (0-9). The size of each image is $16\times 16$.

This experiment follows the setting in ref. \cite{ProCRC}: We randomly select 50, 100, and 300 samples from each digit (class), and use all samples of the testing set for testing. Tables \ref{t_MNIST} and \ref{t_USPS} list the experiment results. We can observe that the proposed LDSR and KLDSR classifiers outperform the other comparison methods. Moreover, the recognition rates of LDSR and KLDSR increase consistently with the increasing number of training image samples. However, NSC has the dropping recognition rate in the same situation.

\begin{table} 
\setlength{\tabcolsep}{10.75pt}
\setlength{\extrarowheight}{3.75pt}
\caption{Recognition rates (RR) of several classifiers on the MNIST database } \label{t_MNIST}
\begin{center}
\begin{tabular}{|l|c|c|c|}
\hline
Classifier & 50 & 100 & 300 \\
\hline\hline
SVM \cite{SVM}     & 89.35 & 92.10 & 94.88\\
NSC \cite{LRC}    & 91.06	& 92.86	& 85.29 \\
CRC \cite{CRC}   & 72.21	& 82.22	& 86.54 \\
SRC \cite{SRC}      & 80.12	& 85.63	& 89.30 \\
KSRC \cite{KSRC}  & 80.32	& 85.86	& 89.88 \\
CROC \cite{CROC}\cite{CROC1}   & 91.06	& 92.86	& 89.93 \\
NDSR \cite{NDSR}  & 91.64	& 94.06	& 95.18 \\
ProCRC \cite{ProCRC}    & 91.84	& 94.00	& 95.48 \\
\hline
LDSR      & 92.62	& 94.73	& 96.03\\
KLDSR      & 92.91	& 94.89	& 96.10\\
\hline
\end{tabular}
\end{center}
\vspace{-0.1in}
\end{table}

\begin{table} 
\setlength{\tabcolsep}{10.75pt}
\setlength{\extrarowheight}{3.75pt}
\caption{Recognition rates (RR) of several classifiers on the USPS database } \label{t_USPS}
\begin{center}
\begin{tabular}{|l|c|c|c|}
\hline
Classifier & 50 & 100 & 300 \\
\hline\hline
SVM \cite{SVM}     & 93.46	& 95.31	& 96.30\\
NSC \cite{LRC}    & 93.48	& 93.25	& 87.85 \\
CRC \cite{CRC}   & 89.89	& 91.67	& 92.79 \\
SRC \cite{SRC}      & 92.58	& 93.99	& 95.86 \\
KSRC \cite{KSRC}  & 92.48	& 93.78	& 95.96 \\
CROC \cite{CROC}\cite{CROC1}   & 93.48	& 93.25	& 91.87 \\
NDSR \cite{NDSR}  & 93.68	& 95.31	& 96.25 \\
ProCRC \cite{ProCRC}    & 93.84	& 95.62	& 96.43 \\
\hline
LDSR      & 94.26	& 96.12	& 96.87 \\
KLDSR      & 94.48	& 96.31	& 96.91 \\
\hline
\end{tabular}
\end{center}
\vspace{-0.1in}

\end{table}

\subsection{Performance on deep learning feature}

This section assesses the proposed classifiers with the deep learning feature on three challenging databases, namely the Oxford 102 Flowers database, Caltech-UCSD Birds (CUB200-2011) database and Caltech-256 database. In the following experiment, VGG-verydeep-19 is employed to extract CNN feature (namely VGG19 features) \cite{VGG19}. The activations of the penultimate layer is used as local features. They are extracted from 5 scales {$2^s, s= -1, -0.5, 0, 0.5, 1$}. We pool all local features together regardless of scales and locations. The final feature dimension of each sample in these databases is 4,096.

\subsubsection{Flowers Recognition}

Oxford 102 Flowers database has 8189 flower images of 102 categories. Different flowers have different scales, pose and lighting conditions. Due to large variations within the category and small difference of different categories, this database is quite challenging and is also widely-used for fine-grained image recognition. We follow the settings in \cite{ProCRC}. Table \ref{t_Oxford102} describes the experiment results. The left part of Table \ref{t_Oxford102} gives the experiment results of the proposed LDSR, KLDSR and several well-known classifiers; the right part of Table \ref{t_Oxford102} lists the results of several state-of-the-art methods for flowers recognition. From Table \ref{t_Oxford102}, we can know that KLDSR obtains the highest recognition rate with the VGG19 features among compared classifiers and the state-of-the-art methods.

\begin{table} 
\setlength{\tabcolsep}{10.75pt}
\setlength{\extrarowheight}{3.75pt}
\caption{Recognition rates (RR) of several methods on the Oxford 102 Flowers database with VGG19-based deep feature } \label{t_Oxford102}
\begin{center}
\begin{tabular}{|l|c|c|c|}
\hline
 &  & State-of-the-art & \\
  Classifier          &  RR  &      Methods & RR \\
\hline\hline
Softmax \cite{Softmax}     & 87.3	& NAC \cite{NAC}	& 95.3\\
SVM \cite{SVM}     & 90.9	& OverFeat \cite{OverFeat}	& 86.8\\
Kernel SVM \cite{SVM}     & 92.2	& GMP \cite{GMP}	& 84.6\\
NSC \cite{LRC}    & 90.1	& DAS \cite{DAS}	& 80.7 \\
CRC \cite{CRC}   & 93.0	& BigCos set \cite{BigCos}	& 79.4 \\
SRC \cite{SRC}      & 93.2	& 	&  \\
KSRC \cite{KSRC}      & 93.3	& 	&  \\
CROC \cite{CROC}\cite{CROC1}   & 93.1	& 	&  \\
ProCRC \cite{ProCRC}    & 94.8	& 	&  \\
\hline
LDSR      & 95.3	& 	 & \\
KLDSR      & 95.7	& 	 & \\
\hline
\end{tabular}
\end{center}
\vspace{-0.1in}

\end{table}

\subsubsection{Birds Recognition}

The Caltech-UCSD Birds (CUB200-2011) database has 11,788 images of 200 bird species. Because bird species are high degree of similarity, this database is quite challenging and is widely-used for fine-grained image recognition. We follow the settings in \cite{CUB200}: Around 30 samples of each bird species are used as training set, the resting images are used as testing set. The experiment results are shown in Table \ref{t_CUB200}. The left part of Table \ref{t_CUB200} lists the experiment results of the proposed LDSR, KLDSR and several well-known classifiers; the right part of Table \ref{t_CUB200} lists the experiment results of several state-of-the-art methods for birds recognition. From Table \ref{t_CUB200}, we can see that KLDSR obtains the highest recognition rate with the VGG19 features among compared classifiers. Moreover, KLDSR obtains the second highest among state-of-the-art methods.

\begin{table} 
\setlength{\tabcolsep}{10.75pt}
\setlength{\extrarowheight}{3.75pt}
\caption{Recognition rates (RR) of several methods on the Caltech-UCSD Birds (CUB200-2011) database with VGG19-based deep feature } \label{t_CUB200}
\begin{center}
\begin{tabular}{|l|c|c|c|}
\hline
 &  & State-of-the-art & \\
  Classifier          &  RR  &      Methods & RR \\
\hline\hline
Softmax \cite{Softmax}     & 72.1	& NAC \cite{NAC}	& 81.0\\
SVM \cite{SVM}     & 75.4	& PN-CNN \cite{PN-CNN}	& 75.7\\
Kernel SVM \cite{SVM}     & 76.6	& FV-CNN \cite{FV-CNN}	& 66.7\\
NSC \cite{LRC}    & 74.5	& POOF \cite{POOF}	& 56.9 \\
CRC \cite{CRC}   & 76.2	& 	&  \\
SRC \cite{SRC}      & 76.0	& 	&  \\
KSRC \cite{KSRC}   & 76.2	& 	&  \\
CROC \cite{CROC}\cite{CROC1}   & 76.2	& 	&  \\
ProCRC \cite{ProCRC}    & 78.3	& 	&  \\
\hline
LDSR      & 79.2	& 	 & \\
KLDSR      & 79.5	& 	 & \\
\hline
\end{tabular}
\end{center}
\vspace{-0.1in}
\end{table}

\subsubsection{Object Recognition}

The Caltech-256 dataset has 256 object classes. Each object class contains at least 80 images. This database has 30,608 images in total. To evaluate the performance of LDSR and KLDSR for object recognition, we follow Ref. \cite{ProCRC}, randomly select 30 images for training, the rest images are used for testing. Table \ref{t_Caltech256} lists the experiment results. All comparison methods are classifiers. As we can observe, LDSR and KLDSR have at least 1.0\% improvements compared to the third-best method (ProCRC). Afterwards, we compare the proposed LDSR and KLDSR with the state-of-the-art methods including four deep learning-based methods. To compare fairly, we follow the common experiment settings: Choose 15, 30, 45 and 60 images from each class for training, respectively, and utilize the rest images for testing. We run 10 times for each partition and show the average recognition rate. The experiments results are listed in Table \ref{t_Caltech256_2}. Observing the results, we see that the proposed classifiers have at least 1\% improvement compared to all state-of-the-art methods.

\begin{table} 
\setlength{\tabcolsep}{10.75pt}
\setlength{\extrarowheight}{3.75pt}
\caption{Recognition rates (RR) of several classifiers on the and Caltech 256 object database with VGG19-based deep feature} \label{t_Caltech256}
\begin{center}
\begin{tabular}{|l|c|}
\hline
~~~~Classifier~~~~ & ~~~~RR~~~~  \\
\hline\hline
Softmax \cite{Softmax}       & 75.3   \\
SVM \cite{SVM}       & 80.1   \\
Kernel SVM \cite{SVM}       & 81.3   \\
NSC \cite{LRC}      & 80.2   \\
CRC \cite{CRC}     & 81.1   \\
SRC \cite{SRC}    & 81.3   \\
KSRC \cite{KSRC}    & 81.7   \\
CROC \cite{CROC}      & 81.7   \\
ProCRC \cite{ProCRC}    & 83.3   \\
\hline
LDSR    & 84.3   \\
KLDSR    & 84.7   \\
\hline
\end{tabular}
\end{center}
\vspace{-0.1in}
\end{table}

\begin{table} 
\setlength{\tabcolsep}{10.75pt}
\setlength{\extrarowheight}{3.75pt}
\caption{Recognition rates (RR) of several state-of-the-art methods on the and Caltech 256 object database with VGG19-based deep feature } \label{t_Caltech256_2}
\begin{center}
\begin{tabular}{|l|c|c|c|c|}
\hline

  Classifier          &  15  &  30 & 45 & 60 \\
\hline\hline
MHMP  \cite{MHMP}     & 40.50	& 48.00	& 51.90	& 55.20\\
IFV \cite{IFV}     & 34.70	& 40.80	& 45.00	& 47.90\\
LLC \cite{LLC}     & 34.36	& 41.19	& 45.31	& 47.68\\
ScSPM \cite{ScSPM}     & 27.73	& 34.02	& 37.46	& 40.14\\
ZF \cite{ZF}     & 65.70	& 70.60	& 72.70	& 74.20\\
LLNMC \cite{LLNMC}     & 68.32	& 71.89	& 74.13	& 75.47\\
LLKNNC \cite{LLNMC}    & 68.55	& 72.09	& 74.07	& 75.36 \\
CNN-S \cite{CNN-S}   & -	& -	& -	& 77.61 \\
VGG19 \cite{VGG19}      & -	& -	& -	& 85.10 \\
NAC \cite{NAC}   & -	& -	& -	& 84.10 \\
ProCRC \cite{ProCRC}    & 80.20	& 83.30	& 84.90	& 86.10 \\
\hline
LDSR      & 81.20	& 84.19	& 86.00	& 87.14\\
KLDSR      & 81.36	& 84.50	& 86.13	& 87.36\\
\hline
\end{tabular}
\end{center}
\vspace{-0.1in}

\end{table}

\subsection{Performance on ImageNet}

The \cite{ProCRC} may have the potential scalability problem when the quite large scale dataset is used [6], such as ImageNet database. They may not be feasible to load millions of samples into memory and may not be easy to solve the inverse of a matrix with millions-dimension. To address this problem, the dictionary learning (DL) techniques are used. We follow \cite{ProCRC}, one simple DL model is used, which is described as
\[\mathop {\min }\limits_{{D_k},{A_k}} ||{X_k} - {D_k}{A_k}||_F^2 + \tau ||{A_F}||_F^2\]
where $\tau $ is a constant. Using the above equation, the dictionary ${D_k}$ will replace the original training set $X_k$ . The dimension of samples of ${D_k}$ is the same as that of $X_k$ . However, the number of samples of ${D_k}$ is different from that of $X_k$ . For example, ImageNet Large Scale Visual Recognition Challenge (ILSVRC) 2012 dataset contains 1.2M+ training images with 1000 classes (around 1300 images per class). Using the above equation, the number of samples of each class is reduced to 50. That is, we have only 50K training images and 50K testing images.\\
In this experiment, the ImageNet database is used. We follow Ref. \cite{ProCRC} and use AlexNet feature extracted by Caffe \cite{Imagenet} (feature dimension is 4096). Table \ref{t_ImageNet} shows the recognition rates of several classifiers with top-1 and top-5 schemes. From Table \ref{t_ImageNet}, we know that KLDSR obtains the best performance with the AlexNet feature on top-5 scheme, and obtains the second-highest performance with the AlexNet feature on top-1 scheme.

\begin{table}
\setlength{\tabcolsep}{10.75pt}
\setlength{\extrarowheight}{3.75pt}
\begin{center}
\caption{Recognition rates (RR) of several classifiers on ImageNet }\vspace{0.05in} \label{t_ImageNet}
\setlength{\tabcolsep}{8.75pt}
 \begin{tabu}{|c|c|c|}
   \hline
Classifier &  Top 5 &  Top 1  \\
 \hline
 \hline

  Softmax \cite{Softmax}      	& 80.4	& 57.4   \\
SVM \cite{SVM}       	& 79.7	& 55.8   \\
NSC \cite{LRC}     	& 77.4	& 53.2   \\
CRC \cite{CRC}     	& 78.5	& 54.3   \\
SRC \cite{SRC}    	& 78.7	& 54.1   \\
KSRC \cite{KSRC}    	& 78.8	& 54.3   \\
CROC \cite{CROC}      	& 78.8	& 54.4   \\
ProCRC \cite{ProCRC}   	& 80.1	& 56.3   \\
\hline
LDSR    	& 80.7	& 56.8   \\
KLDSR    	& 80.9	& 57.1   \\
  \hline
  \end{tabu}
\end{center}
\label{t_ilids}
\end{table}

\subsection{Evaluation of the effect of $s$}
\begin{figure}[t]
\begin{center}
\includegraphics[width=0.9\linewidth]{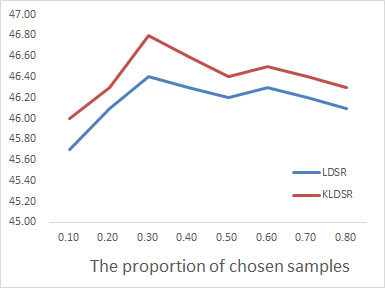}
\end{center}
\vspace{-0.1in}
   \caption{Evaluate the effect of the proportion of chosen samples.} \label{fig_K}
\end{figure}

We need select $s$ samples to constitute the locality-based training set $Y$. This section evaluates the effect of $s$. The LFW face database is utilized in this experiment. The setting is the same as that of Section A. $s$ is set as {$0.1, 0.2, ..., 0.8$} times of the number of entries samples of all classes. Figure \ref{fig_K} shows the experiment results. As we can see, the proposed methods obtains good performance when the number of chosen samples $s$ is moderate. The best performance can be obtained when the proportion of chosen samples belongs to [0.2,0.5].

\section{Conclusion}

In this paper, we have proposed a discriminant-based sparse optimization learning model. Based on this model, locality-based discriminant-based sparse representation (LDSR) has been proposed for vision recognition. LDSR obtains the good classification performance because discriminant can help to increase the difference of samples in different classes and to decrease the difference of samples within the same class. Moreover, kernel locality-based discriminant-based sparse representation (KLDSR) was further proposed based on the kernel techniques. KLDSR can take advantages of kernel techniques, discriminant and sparse. To demonstrate the performance of proposed classifiers, extensive experiments have been carried out on several databases including the million-level database: ImageNet. All experiment results prove the effectiveness of the proposed LDSR and KLDSR.

%

{\small
\bibliographystyle{IEEEtran}
\bibliography{DSRbib}
}

%
%

\end{document}